\documentclass[acmsmall]{acmart}
\AtBeginDocument{%
  }

\setcopyright{acmlicensed}
\copyrightyear{2026}
\acmYear{2026}
\acmDOI{XXXXXXX.XXXXXXX}

\acmJournal{JACM}
\acmVolume{37}
\acmNumber{4}
\acmArticle{111}
\acmMonth{8}

\usepackage{booktabs}
\usepackage{longtable}

\begin{document}

\title{Modeling Social Dynamics with an LLM-Enabled Agent Based Network-Dynamic (LAND) Model}

\author{Lynnette Hui Xian Ng}
\affiliation{%
  \institution{Carnegie Mellon University}
  \city{Pittsburgh}
  \country{USA}}
\email{lynnetteng@cmu.edu}

\author{Kathleen M. Carley}
\affiliation{%
  \institution{Carnegie Mellon University}
  \city{Pittsburgh}
  \country{USA}}
  \email{carley@andrew.cmu.edu}

\renewcommand{\shortauthors}{Ng \& Carley}

\begin{abstract}
Social dynamics encode the process in which individual network and discourse interactions aggregate into collective influence, narrative dominance and coordinate behavior.
This paper uses the the GhostField architecture, a  hybrid LLM-Enabled Agent Based Network-Dynamic (LAND) model as a social simulation framework to build the AuraSight scenario. In the AuraSight scenario, 314,244 heterogeneous cyber social agents and human actors exchange 529,327 messages over 30 days surrounding a fictional international song-writing contest. We methodologically examine emergent social dynamics across four analytical layers: ego-network topology, semantic network evolution, coordination dynamics and influence dynamics. Our results show how generated social simulations do also produce social dynamics, and how the dynamics of coordination and influence emerge not from individual agents but from the recursive interaction between network topology and narrative exchange.
\end{abstract}



\received{}
\received[revised]{}
\received[accepted]{}

\maketitle

\section{Introduction}
Understanding social dynamics of coordination and influence is a central research thrust in social network research. Foundational theories of social influence argue that individual beliefs and behaviors emerge through interpersonal influence processes within social structures~\citep{friedkin1990social}.
When studying such dynamics on social media platforms, observational studies of user behavior is often employed to profile key events like elections and crises. However, while observational approaches provide valuable empirical insights into emergent social dynamics, it is often difficult for analysts to conduct ethical and controlled that allows the exploration of counterfactual scenarios in such social systems~\citep{ng2022online,graham2024coordination,carley2020social}. 

Simulation approaches are often employed for the study of emergent social dynamics and for intervention testing~\citep{fan2018agent,ng2026botsim}. Traditional Agent-Based Models (ABMs) rely on simplified communication which allows for scale and interpretability, but lack the linguistic realism of social media discourse~\citep{adornetto2025generative,ng2026botsim}. Recent advances in Large Language Models (LLMs) can generate realistic language conversations between agents which can be seeded with persona and memory variables but existing approaches often lack the network-constrained interaction dynamics necessary to reproduce emergent social behaviors observed in large-scale online ecosystems~\citep{larooij2025large}. This study therefore adopts a hybrid LLM-Enabled Agent Based Modeling approach to simulate and study the social dynamics of social media agents, combining the scalability of ABMs with the conversational realism of LLMs.

We use the GhostField simulation framework~\citep{carley2026ghostfield} that integrates LLM-generated conversational threads into an ABM modeling framework to create a fictitious scenario called AuraSight. This scenario simulates the propagation of narratives on the social media platform X during an international song-writing contest. The simulated agent-agent conversations and interactions are examined across several layers of emergent social dynamics: changes in network topology, linguistic patterns, coordination dynamics and influence dynamics. With our methodology and results, we make the following contributions:


\begin{itemize}
    \item \textbf{Methodologically, } we use the GhostField framework to develop the AuraSight scenario which incorporates conversational realism of LLMs into the interaction fidelity of ABMs within the context of network dynamics, building a LLM-Enabled Agent Based Network-Dynamic (LAND) model architecture.
    With our LAND model architecture, we generated 314,244 agents, 529,327 unique messages and their corresponding interactions across a 30-day time frame that follows a few key events. This data is generated in the form of a X V1 API JSON format, thus networks can be extracted from the data because there is both message content and interaction metadata.
    
    \item \textbf{Empirically, } we establish a methodology for studying social dynamics. This allows us to study the coordination and linguistic patterns for data generated from an LLM-Enabled Agent Based Network-Dynamic (LAND) model.
    
    \item \textbf{Theoretically, } we demonstrate that hybrid LLM-enabled ABMs provide a computational framework for studying socially embedded agents whose behaviors are fundamentally interaction-driven. We establish that in social media, the social dynamics of coordination and influence emerge not from individual agent cognition but from the recursive interaction between network topology and narrative exchange; and that the social simulation we built is comparable to real-life dynamics through several stylized facts.
\end{itemize}

This paper begins with a review of social simulation literature in \autoref{sec:literature}, then describe the Ghostfield architecture and the AuraSight scenario in \autoref{sec:architecture}. We examine the social dynamics from the simulated data in \autoref{sec:dynamics} in terms of ego-network evolution, semantic network change, agent-agent coordination dynamics and influence maneuver dynamics. In \autoref{sec:stylized_facts}, we show that the social dynamics generated from the simulated data are in line with observed social dynamics from empirical data through stylized facts.
Finally, we discuss the results and future work for the hybrid LAND architecture in \autoref{sec:discussion}, before concluding in \autoref{sec:conclusions}.

\section{Literature Review}
\label{sec:literature}
Modeling and simulating social interactions provide a foundation for understanding collective phenomena. Social simulation is especially useful in environments where direct experimentation is infeasible. Traditionally, social simulation used Agent-Based Models (ABMs) which emphasizes structural and behavioral rules that govern interactions. Classic work like the the Friedkin-Johnson social influence model argues that individual beliefs and behaviors are shaped by both intrinsic values and interpersonal influences from social relations~\citep{friedkin1990social}. Computational models like Construct~\citep{carley2009etiology,carley1990group} extend this logic by modeling how agents exchange information through social and knowledge networks, showing how local interaction processes can produce group-level and organizational-level outcomes~\citep{carley2009etiology,carley1990group}. Most recently, the development of Large Language Models (LLMs) allows for the integration of generative AI to construct linguistically rich simulations for better conversational realism~\citep{ng2025llm,taillandier2025integrating}. 

Agent-Based Models have been used to explain viral information spread and cascade behaviors in online networks, and network evolution dynamics that include the formation and breakage based on homophily and social influence mechanisms and of polarized clusters and echo chambers~\citep{betts2022effect,liu2023emergence}.
An ABM simulates successive agent-agent and agent-environment interactions across time, allowing for the observation of emergent behaviors that connect micro-level individual agent behavior to macro-level community patterns~\citep{betts2022effect}. Even small perturbations in the information environment can shift group cognition and organizational-level consensus through iterated local interactions~\citep{carley2009etiology}. For example, X feeds modeled as a discrete event simulation can aid studies of the emergent behavior of two bot-based disinformation maneuvers, bridging and backing, which revealed that bots are only effective when correctly embedded in the network~\citep{beskow2019agent}. 
However, traditional ABMs face limitations in modeling social media discourse because agent communications are represented by a vector of numbers, in binary states or numerical beliefs or stylized signals rather than the language rich messages that characterize social media discourse~\citep{adornetto2025generative,ng2025llm}. This limits their ability to model cognition, discourse networks and the content-level mechanisms through which social dynamics unfold online.


Turn taking dynamics are central to conversational social network analysis, because they shape the sequential structure of online discourse: who responds to whom, in what order, and which narratives gain traction. From a social networks perspective, the pattern of turn-taking creates directed interaction chains that privilege central, high-degree actors and amplify the narrative reach of the actors~\citep{gibson2005taking,zhao2011social}.
Pure ABM agents do not produce or interpret language, and therefore cannot fully simulate conversational discourse. 
This gap motivates the integration of LLMs into social simulations to endow agents with expressive, goal-directed communication capabilities. Advances in LLM-based simulation have enabled more realistic agent communication and behavior synthesis. These social simulations range from dialog-driven settings~\citep{feng2023towards} (e.g., social interaction, question-answering and game-based) to task driven settings (e.g. software-development, code-testing)~\citep{mou2024individual}. 
Such simulations have crafted artificial societies with LLM-powered agents that are capable of autonomous memory and planning, and demonstrate emergent social behaviors~\citep{park2023generative}. For example, OASIS is a social simulation that models the information propagation dynamics in X and the herd effect in Reddit with a total of one million agents ~\citep{yang2024oasis}. However, many existing LLM-based simulations lack the explicit modeling of social structure, role differentiation and network-level interactions. Agents often operate in sandboxed contexts without persistent relationships or inter-agent dependencies that mirror real-world social dynamics. Our work builds upon these foundations by building a social simulation of a community's response to events, where agents are endowed with personalities and language abilities from an LLM, and interact in an ABM define the structure in which interactions occur. This results in a social dynamic where the structure and discourse co-evolve in harmony.

Beyond simply building a social simulation, one must also analyze the social dynamics emergent from such simulated environments. Social dynamics describes how social systems and actors change over time~\citep{tuma1984social}, and has been studied through several complementary lenses. At the structural level, degree distribution, centrality metrics and preferential attachment explain why a small number of agents can accrue a disproportionately large amount of attention~\citep{jeong2003measuring,azadbakht2017distributed}. At the community level, community detection and clustering coefficients capture the formation of topically homogeneous echo chambers~\citep{cinelli2021echo} and drastically polarized opinion groups~\citep{liu2023emergence}. At the content level, semantic network analysis of co-occurring linguistic tokens like hashtags reveals how narratives cluster and topic saliencies ~\citep{diesner2005revealing,alieva2026dynamics}. Our work builds on this foundation by constructing a social simulation from a storyline, where the agents are embedded with personality and language abilities from LLMs and the operational strategies from ABMs. The storyline constraints the social dynamics, which can be analyzed from both structural and discursive perspectives. 

\section{LLM-Enabled Agent Based Network-Dynamic (LAND) Model}
\label{sec:architecture}
Our LAND model is implemented using the GhostField architecture which provides the computational scaffolding for our simulation. We then use the scaffolding to develop the AuraSight scenario which instantiates the architecture as a concrete social environment. The architecture defines the interaction grammar of the LAND model, while the scenario instantiates the actors, topics, narratives and event timeline.
Populating the GhostField framework specifies the time-zero conditions. Although the system is stochastic, each run produces a slightly different realization of agent communications, but crucially, the structural and behavioral constraints ensure that the macro-level social dynamics remain reproducible over time.



\subsection{The GhostField Architecture}
\label{sec:ghostfield}
GhostField is a system of systems used to generate the data for a social simulation of communities of agents interacting online around a storyline, or a sequence of events~\citep{carley2026ghostfield}. GhostField has two major subsystems: (1) AESOP which supports the scenario design, and (2) SynTelX which takes the generated scenario template from AESOP and generates the communications and messages among the agents. The GhostField system simulates the propagation of narratives by combining the structured models of classical ABMs with the linguistic expressions of LLMs. The architecture has two parts, which is briefly explained in this section. We urge the reader to read the original paper for details.

AESOP is the scenario generator which designs a scenario based on a storyline, written through a sequence of events. The storyline constraints the alignment of the social and semantic networks. The scenario is parameterized with the following: (1) a set of time-stamped events; (2) a set of topics, each with a pro- and anti- narrative stand, and which topics are active at which time; (3) a roster of agents (humans, cyber social agents, dredgers) assigned to topics, stances towards topics, and narratives that they will engage in, and their non-evolving agent attributes that influence behavior; (4) a set of behaviors that the agents will undertake under specified conditions, (5) narrative seeds for each topic; and (6) the duration over which the topics are active. This parameterization is analogous to initializing the prior belief distribution in network adjacency matrices in classical computational social influence models~\citep{friedkin1990social,carley2009etiology}. 
The output of the AESOP system is a scenario template in a JSON format that contains all the necessary parameters to run the simulation, like detailed agent persona description, topic description and event sequences. 

The SynTelX system~\citep{hicks2024ai} takes the JSON output from the AESOP system as an initial simulation input, and generates interactions and communication messages. The parameters from the AESOP scenario specifies the conditions under which agents sends what messages about what issues, and the conditions under which agents will respond to messages on certain issues. 
These heuristics provide a scaffolding against the actual specific messages are created by SynTelX. Every time SynTelX is run, one will get the same scaffolding but a new set of messages where different sets of actors may be acting.

This simulation generation step has three main steps: agent activation, agent action and generation of linguistic text. Agent activation follows a probabilistic behavioral rule that is drawn from a geometric distribution to reflect the stochastic posting rhythm of social agents~\citep{zhou2023circadian}. An active agent contributes content to the simulation, whether it is through an original post, retweet, or post reply. Each agent persona has a pre-defined action set. For example, a content generation agent can only create original posts and an amplifier agent can only retweet. If an agent is active, the corresponding text is then generated. The agent first selects the topic to respond to, which is a probablistic function dependent on the agent's intrinsic preferences and the topic's popularity, and the text generation is performed using an LLM conditioned on the agent's persona, the topic selected and other persona parameters.
The resultant output from SynTelX is a corpus of messages for each day of the overall scenario as a JSON file. In this study, we specify the JSON file to look like tweets collected using the Twitter V1 API, which allows us to form social interaction networks (who retweets, mentions, quotes whom), and semantic networks (which ideas and hashtags co-occur together). We are then able to use the interaction and communication traces of the agents with social network methods for analysis.


\subsection{The AuraSight Scenario} 
\label{sec:aurasight}
\begin{figure}
    \centering
    \includegraphics[width=\linewidth]{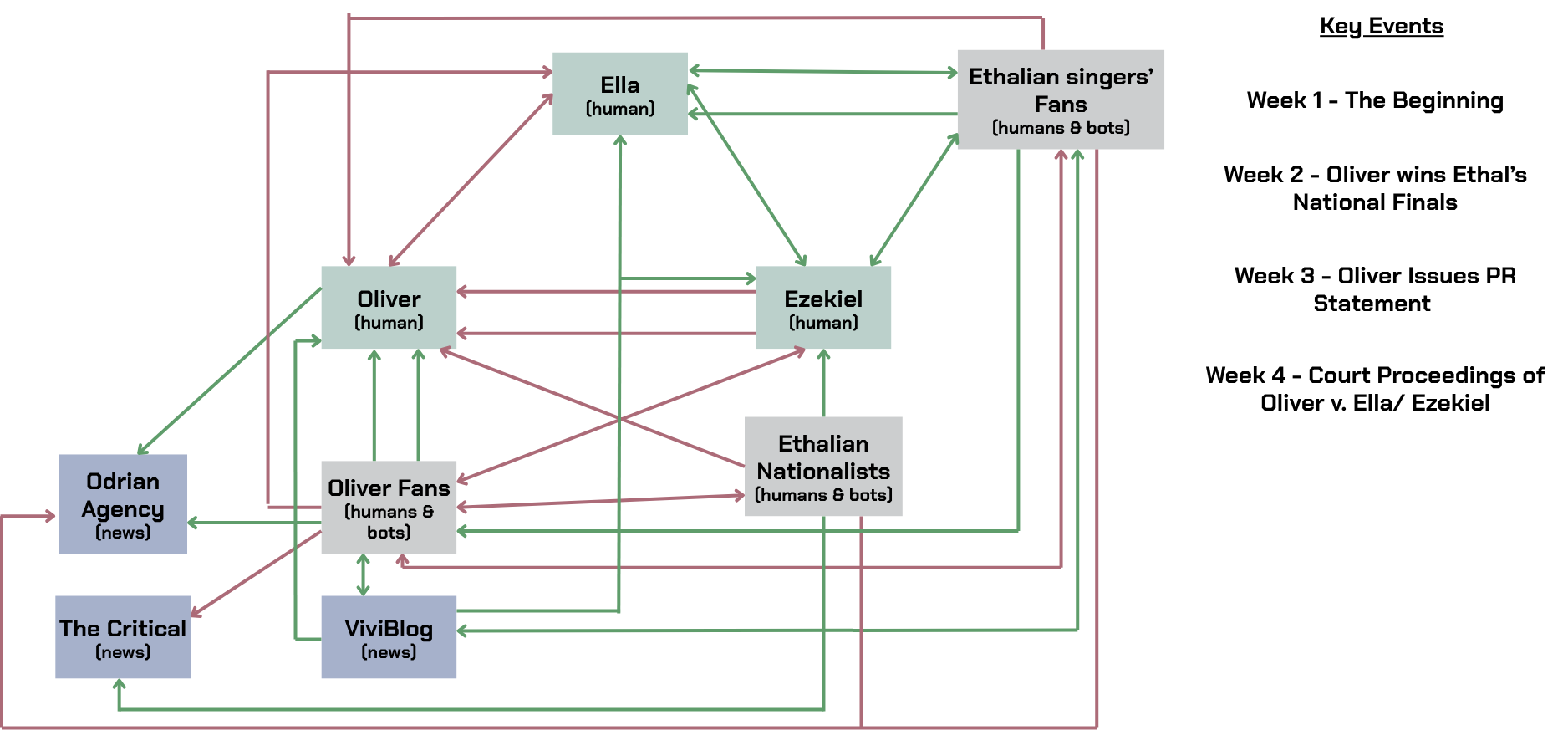}
    \caption{AuraSight support diagram of the key actors, support factions and news agencies. Green arrows from A -> B indicate that A supports B, and red arrows from A -> B indicate that A dislikes B. }
    \label{fig:stance_diagram}
\end{figure}

The AuraSight scenario instantiates the GhostField architecture for a concrete set of events and communities. This scenario simulates the online banter on X around the fictional AuraSight singing competition, which is loosely based on EuroVision competitions. The 2030 AuraSight competition is set to be hosted in Nareth. The scenario takes place over four weeks, where each week has a series of key events. Week 1 begins the scenario where three main competitors (Oliver, Ella, Ezekiel) are actively campaigning for support and publicity.
In Week 2, Oliver, an Odrian-born singer, enters and wins Ethal's national finals and will represent Ethal on the international stage despite being a citizen of Odria. Odria had once conquered Ethal, so the two countries have a fraught 200-year history. Oliver's win divides public opinion. Supporters see it as a cultural reconciliation while nationalists demand disqualification and pursue legal action. 
In Week 3, Oliver delivers a PR statement that argues that the two nations are brothers, adding a personal dimension to the political rift. 
In Week 4, Ella and Ezekiel team up to fight a legal battle against Oliver's win. 
Over the four week course of the scenario, competing factions of fans (Oliver vs Ella/Ezekiel), journalists and bot networks shape the public narrative across social media. \autoref{fig:stance_diagram} presents a stance diagram of the key actors, support factions and news agencies.

In the AuraSight scenario, we have three main types of agents. First, we have human agents that represent real humans. 
These human agents are heterogeneous and each has different sets of issues that they are interested in.
Second, we have cyber social agents, which are archetypes of automated bots. Instead of just a generic bot agent type, we bring the simulation closer to reality by instantiating fifteen archetypes of bots that we derive from past work~\citep{ng202635,ng2026bots}, each with their own behavior characteristics. For example, a news agent is primarily responsible for posting news or news aggregates, and a bridging agent straddles between conversational groups. In our analysis, while these agents have different behavioral archetypes, we will collectively call them ``bot agents", because we are not studying the nuanced differentiation in the social dynamics as affected by the different types of bots. Third, we have the dredger agents who can be bot-like or human-like agents, and whose primary purpose is to latch on to conversations to promote unreliable websites and narratives.
Each agent has their unique behavioral roles and personas, specifically, their topics of interest and stances towards topics. The behavioral constraints of each class of agents are derived from empirically observed online behaviors. 
These agents are first manually instantiated by specifically defining the necessary values for each attribute. After the key actors and 50 other agents were instantiated, we used the GPT-4.1 Large Language Model to generate another 50 agent personas that are based on the manually created personas, which enables us to create agents at scale.

Using AESOP, this storyline is converted into a series of parameters which is then fed as an input to SynTelX. SynTelX then generates all the community discussions along the key events of the four weeks by the instantiated agents. Interactions between the agents are governed by network science principles like preferential attachment and key influencers. The network structure, agent behavior and language generation operate jointly within a simulated social environment. \autoref{tab:aurasight_data} presents the information about the generated network data using the GRAND network reporting guidelines \citep{nealguidelines}.

\begin{table}[ht]
\centering
\small
\begin{tabular}{lr}
\toprule
Date of event & 2030-05-13 to 2030-06-13 \\
Number of Nodes & 314,244 \\
Degree distribution & 5.495e-08$\pm$3.235e-06 \\
Number of Edges & 245,801 \\
Direction of Edges & Undirected (for simplicity of analysis) \\
Edge weight distribution & 1.17$\pm$1.01 \\
\bottomrule
\end{tabular}
\caption{AuraSight data parameters, presented in the GRAND network data reporting guidelines \citep{nealguidelines}.}
\label{tab:aurasight_data}
\end{table}

\section{Examining Social Dynamics}
\label{sec:dynamics}
This simulation produces four weeks of simulated X data around the event AuraSight. The simulated AuraSight event is then examined for the agent network topologies and the linguistic patterns of narrative exchange which results in models of social dynamics of coordination and influence. This examination is done via \textbf{temporal social dynamics}, where the social dynamics are examined for their change across four key events in the storyline. These four key events lay out over four weeks: The Beginning (Week 1), Oliver Wins Ethal's National Finals (Week 2), Oliver Issues Statement (Week 3), and Court Proceedings of Oliver v. Ella/Ezekiel (Week 4). Each week thus presents an analysis time window.
This section describes the methodology and the subsequent results used to profile each of the four social dynamics. 



\subsection{Agent Ego-Network Topologies}
At the network structure level, the interaction patterns generated by our simulation architecture gives rise to distinct ego-network topology of agents. Ego-network evolution has been identified as a sensitive indicator of agent salience in dynamic social systems~\citep{cekini2026impact}. Temporal changes in ego-network size and density thus reflects shifts in an agent's information access and influence capacity as the narrative environment evolves.

\begin{figure}
    \centering
    \includegraphics[width=\linewidth]{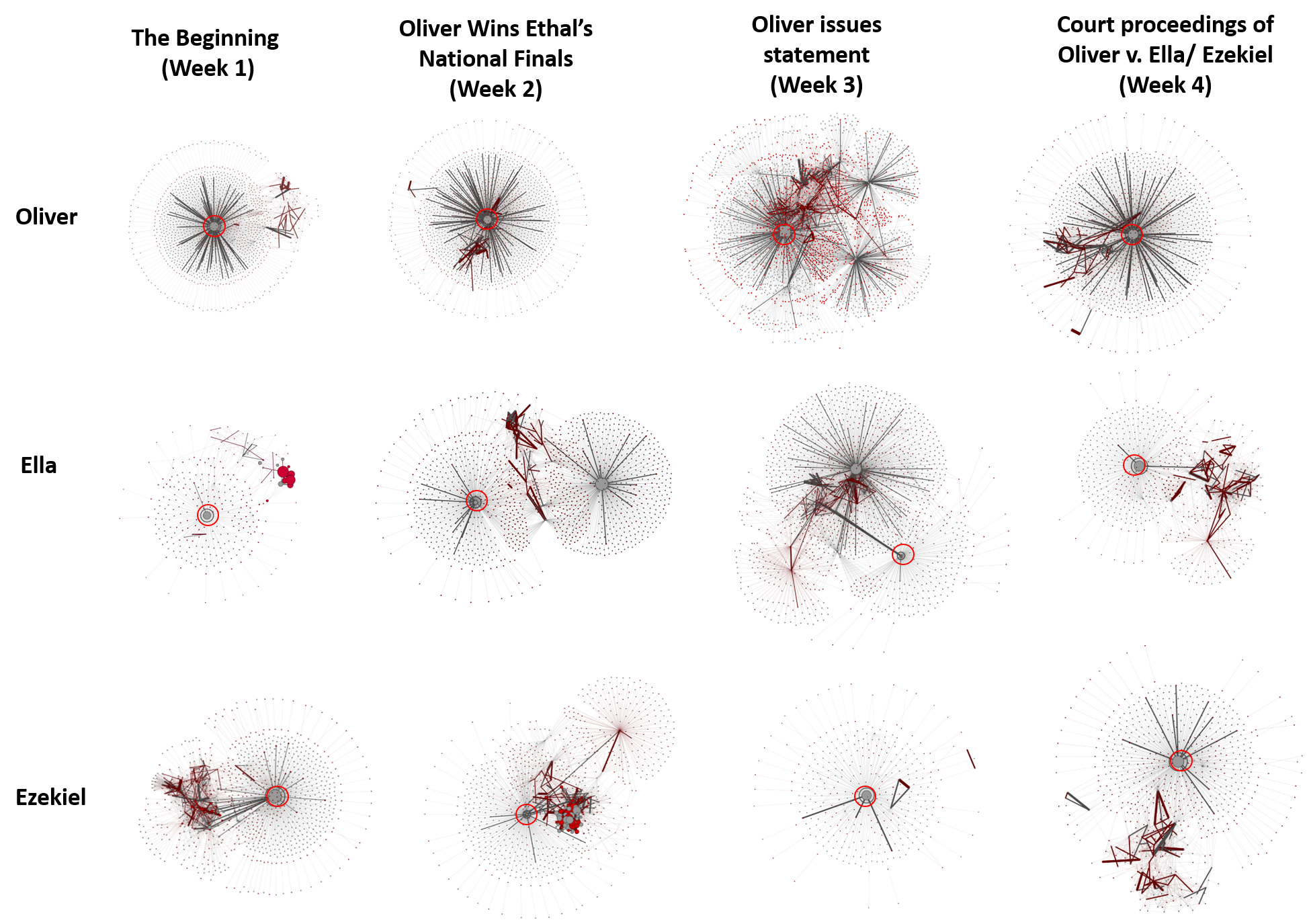}
    \caption{Two-hop Ego Networks of key actors over Time. Nodes represent actors, which are sized by eigenvector centrality. Edges between nodes represent any interactions between the two actors. Red nodes represent bot agents, gray nodes represent human agents.}
    \label{fig:ego_networks}
\end{figure}

To study ego-network evolution, we constructed ego-network graphs as all-communication networks. All-communication network graphs $G_e=(V_e,E_e)$ consists of users as nodes $v_i\in V_e$ and edges $(e_1,e_2) \in E_e$ indicates that user $v_1$ and user $v_2$ had interacted with each other via any of the social media affordances (retweet, reply, @mention). The evolution of ego-networks of the key actors across pivotal events allows the assessment of the impact of content-driven behaviors. For studying temporal social dynamics, we generate the two-degree ego networks $G_e$ for the key protagonists of the AuraSight story, Oliver, Ella and Ezekiel for each event. The two-hop ego-network would contain the immediate connections of the key actors and the connections of the immediate connections. \autoref{fig:ego_networks} presents the network graphs of the ego-networks over time, and \autoref{fig:main_character_centrality} in \autoref{sec:appendix_egonetwork} presents a detailed plot of eigenvector and betweenness centrality values of the key actors.
The betweenness centrality of the key actors are high in Week 1 as the different campaigns are building up support for their person for the contest. The betweenness centrality drops as the scenario progresses as the intense campaigning and publicity period fades and the conversation organically revolves around the storyline.
Overall, we observe shifts in network topologies as a function of the change in events, such as groups fissioning off as Oliver wins the finals and polarized groups of users as the geo-political debate around Oliver heightens and factions form around the legal battle.

Oliver exhibits a consistently dominant structural position across all four snapshots. He also maintains a high eigenvector centrality within his ego-network (0.703 at Week 1, declining modestly to 0.697 at Week 4), and extremely high betweenness centrality (0.988 at Week 1, 0.981 at Week 4). He is the most connected node in his ego-network, regardless of the events that are unfolding. Oliver's centrality scores remain rather stable despite a narrative environment that shifts from a competitive one to an adversarial one against him, is consistent with the Matthew effect in social networks~\citep{liu2024opinion,rigney2010matthew}. The structural prominence the agent accrues through the early events proves robust to subsequent narratives; i.e., the rich get richer. 
Visually, Oliver's ego network transitions from a moderately dense star topology at Week 1 (The Beginning) to a bot-heavy network with multiple clusters at Week 3 (Oliver issues statement), before contracting again at Week 4 (Court Proceedings). Such reflects the behavioral scripting of bot archetypes that amplify the narratives of high-centrality human actors, and the spawning off of multiple viewpoints when an event such as a statement is issued, consistent with empirical observations of bot amplification behavior in real-world social media~\citep{ng2025global}. 

Ella and Ezekiel both begin with sparse ego-networks at Week 1, where their eigenvector centrality values are lower compared to Oliver (Ella 0.05, Ezekiel 0.6, Oliver 0.70). Ezekiel is rather dominant in his own network (as measured by centrality measures), while Ella is peripheral in her own network until Week 4, where her eigenvector centrality jumps sharply from less than 0.2 to 0.58.
Ezekiel’s eigenvector centrality peaks sharply at Week 3 (Oliver Issues Statement) before declining at Week 4, while Ella’s centrality is most prominent at Week 4 (Court Proceedings), reflecting the sequential escalation in which each antagonist becomes structurally salient at the point when the narrative most directly implicates them. This staggered centrality accumulation mirrors the dynamics of reactive network growth documented in empirical event-driven social media studies, where peripheral actors accrue ties only when the narrative frame shifts to render them relevant~\citep{cauteruccio2026structure,granovetter1985economic}.

\begin{figure}
    \centering
    \includegraphics[width=\linewidth]{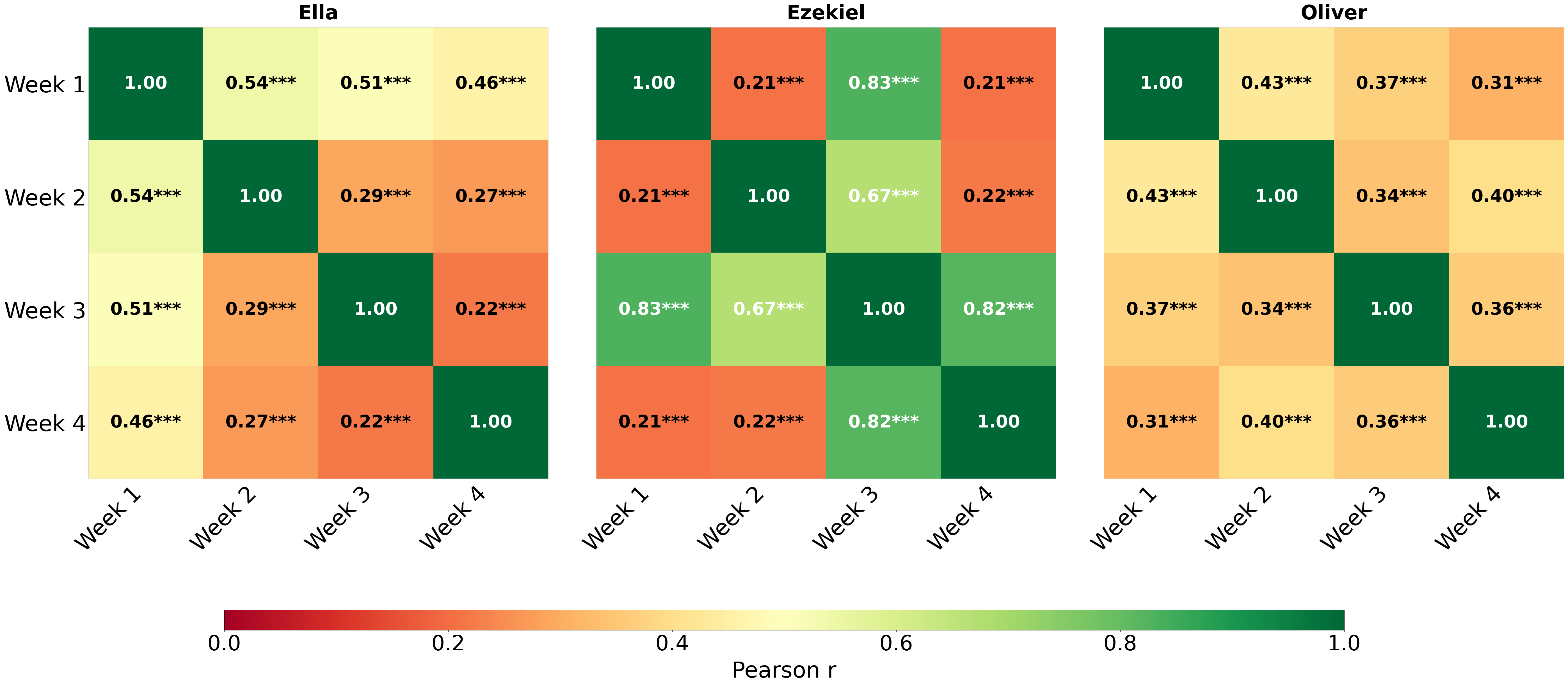}
    \caption{QAP heatmap comparison of ego-networks of key actors across events. ** indicates significant difference at the $p<0.01$ level, *** indicates significant difference at the $p<0.001$ level.}
    \label{fig:ego_qap_heatmap}
\end{figure}

We also used the QAP (Quadratic Assignment Procedure) test on pairwise ego-networks to identify whether the ego-network pairs have correlated structure beyond chance. \autoref{fig:ego_qap_heatmap} presents the results of the correlation, with the detailed statistics in \autoref{sec:appendix_semantic}. The QAP comparisons show that, 
In general, the correlation between sequential weeks tend to be higher correlations than the weeks further apart (i.e., Week 1 is more correlated with Week 2 than it is with Week 4). For Oliver, his Week 1-Week 2 correlation is $r=0.43$, while Week 1-Week 4 correlation is $r=0.31$. For Ella, her Week 1-Week 2 correlation is $r=0.54$, while her Week 1-Week 3 correlation is $r=0.22$. These patterns indicate progressive structural divergence from the initial ego-network initialized as the events accumulate, resulting in changing interactions.
Ezekiel, though, presents different correlation patterns. His Week 1-Week 2 correlation is $r=0.22$ then Week 2-Week 3 at $r=0.67$, before finally being notably high at Week 3-Week 4 ($r=0.82$). This difference in pattern can be attributed to Ezekiel being upset about Oliver's win and consciously been campaigning in his networks.

These observed structural differences of the three key actors emerge from different agent-specific interaction rules and patterns, based off their seeded stories and personas, rather than explicit definitions of centrality outcomes. By including LLM in our agent-based dynamic network model, not only can we create heterogeneous actors, but also the actors are heterogeneous in over-time behavior~\citep{park2023generative}, which we observe in our generated data. In contrast, should using a homogeneous interaction model for each agent like preferential attachment of the network would mean that the patterns of ego-network interactions would look similar for all agents across time~\citep{barabasi1999emergence}.
Such mirrors recognizable social network configurations where the networked influence of actors change organically as the discourse evolves~\citep{boyd2010social}.


\subsection{Linguistic Patterns of Narrative Exchange}
To study linguistic patterns, we construct semantic networks. Semantic network analysis has been widely employed to map the discursive landscape of online political events~\citep{alieva2026dynamics,diesner2005revealing,danaditya2022curious}. Co-occurring linguistic tokens like hashtags are typically used as conceptual proximity and for topic clustering. A denser, more inter-connected semantic network often indicates a convergence of discourse around a unified narrative frame, while a fragmented semantic network with weakly connected components signals topic diversification or narrative polarization~\citep{danowski1993network,bail2016combining}.

\begin{figure}
    \centering
    \includegraphics[width=0.75\linewidth]{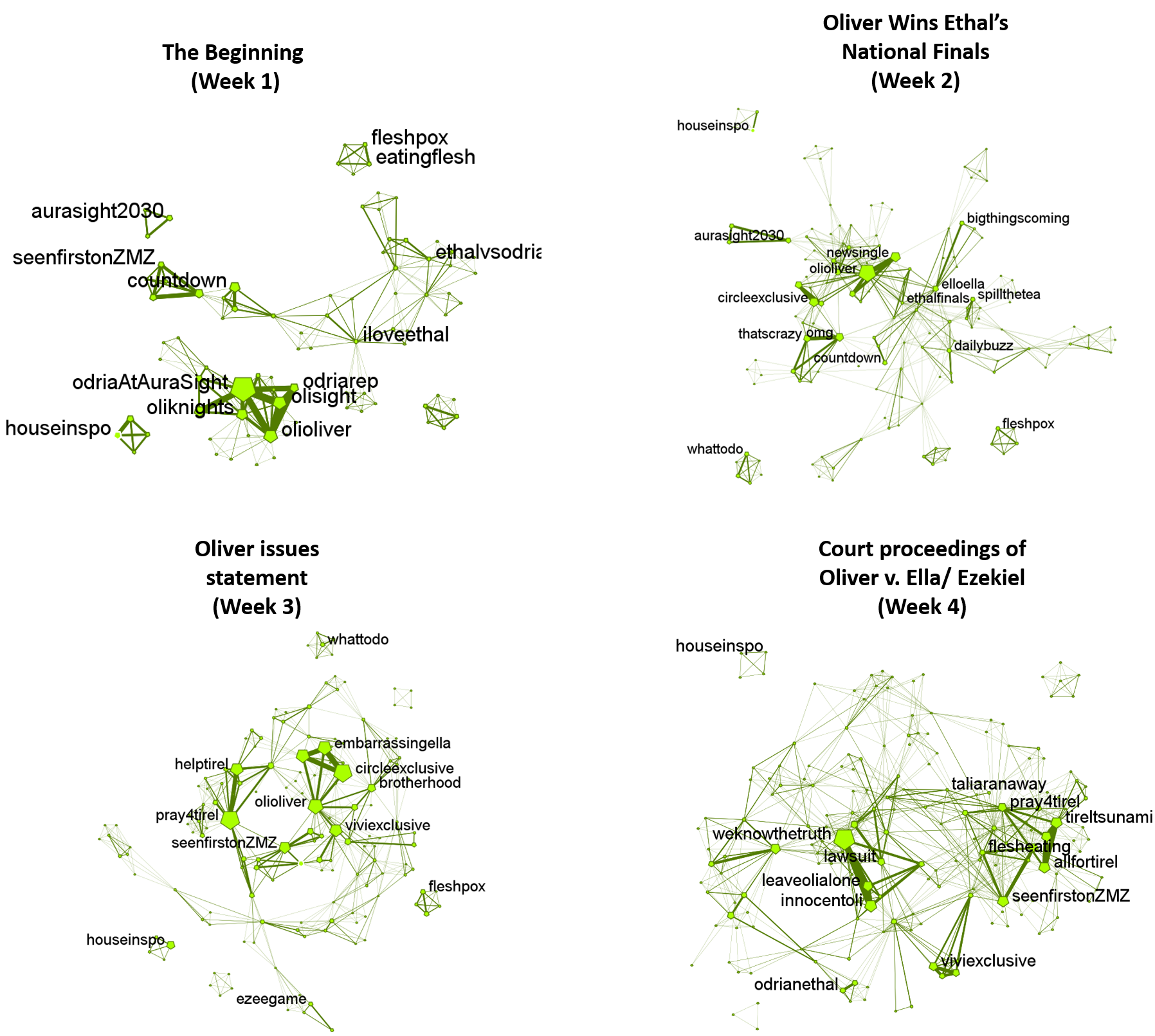}
    \caption{Semantic Networks. Nodes are hashtags, edges represent co-occurrence of hashtags in the same tweet. Nodes are sized by usage count.}
    \label{fig:semantic_networks}
\end{figure}

For studying temporal patterns of narrative exchange, we construct semantic networks at each event step. In the semantic network graph $G_s=(V_s, E_s)$. Nodes $v_i\in V_s$ represent hashtags and edges $(e_1,e_2)\in E_s$ represent the co-occurrence of the two hashtags $e_1$ and $e_2$ in the same tweet. We constructed one $G_s$ for each event, then pruned the network down to the network backbone using the Brandes algorithm~\citep{brandes2001faster}.
Semantic networks are extremely dense compared to social networks, so paring we pared the network down to the backbone to be able to better analyze the core hashtag interactions. The semantic network graphs over the four event points are presented in \autoref{fig:semantic_networks}.

The semantic network evolution reveal a consistent pattern of network growth across the four temporal snapshots. The network expands from 85 nodes and 238 edges at Week 1 to 181 nodes and 730 edges at Week 4, reflecting a cumulative broadening of the discourse vocabulary as the narrative conflict escalates. Modularity scores remain high throughout (0.743 to 0.779), indicating persistent community structure in which hashtags cluster into topically coherent groups rather than separate discourse. The number of communities peaks at Week 2 (15 communities), coinciding with the surge of attention around Oliver’s National Finals win, before stabilizing at 13–14 communities in subsequent snapshots as competing narrative frames consolidate. 

In terms of network content, the hashtag network at Day 1 consists of two broad clsuters, reflecting the pro-Oliver (\textit{\#iloveethal, \#olioliver, \#odriarep}) and anti-Oliver (\textit{\#ethalvsodria, \#seenfirstonZMZ}) narrative strands. 
By Week 2, the network becomes substantially denser (Week 2: 491 edges vs Week 1: 238 edges), and the largest connected component expands from 44 to 126 nodes, reflecting narrative convergence around the pivotal competition event. At Week 3 where Oliver issues a statement, the network shows structural reorganization. The competition hashtags recede, and more emotional hashtags surface (\textit{\#brotherhood,\#circleexclusive}), consistent with how framing theory accounts that personal narratives can restructure topic salience~\citep{entman1993framing}. At Week 4, the court proceedings ensue and the network reached its largest configuration (730 edges, largest component 168 node), but it fractures into legal and conflict-framing clusters (\textit{\#lawsuit, \#innocentoliver, \#leaveollialone, \#wewantthetruth}), signalling a narrative transition from cultural controversy to an adversarial legal conflict.

\begin{figure}
    \centering
    \includegraphics[width=0.5\linewidth]{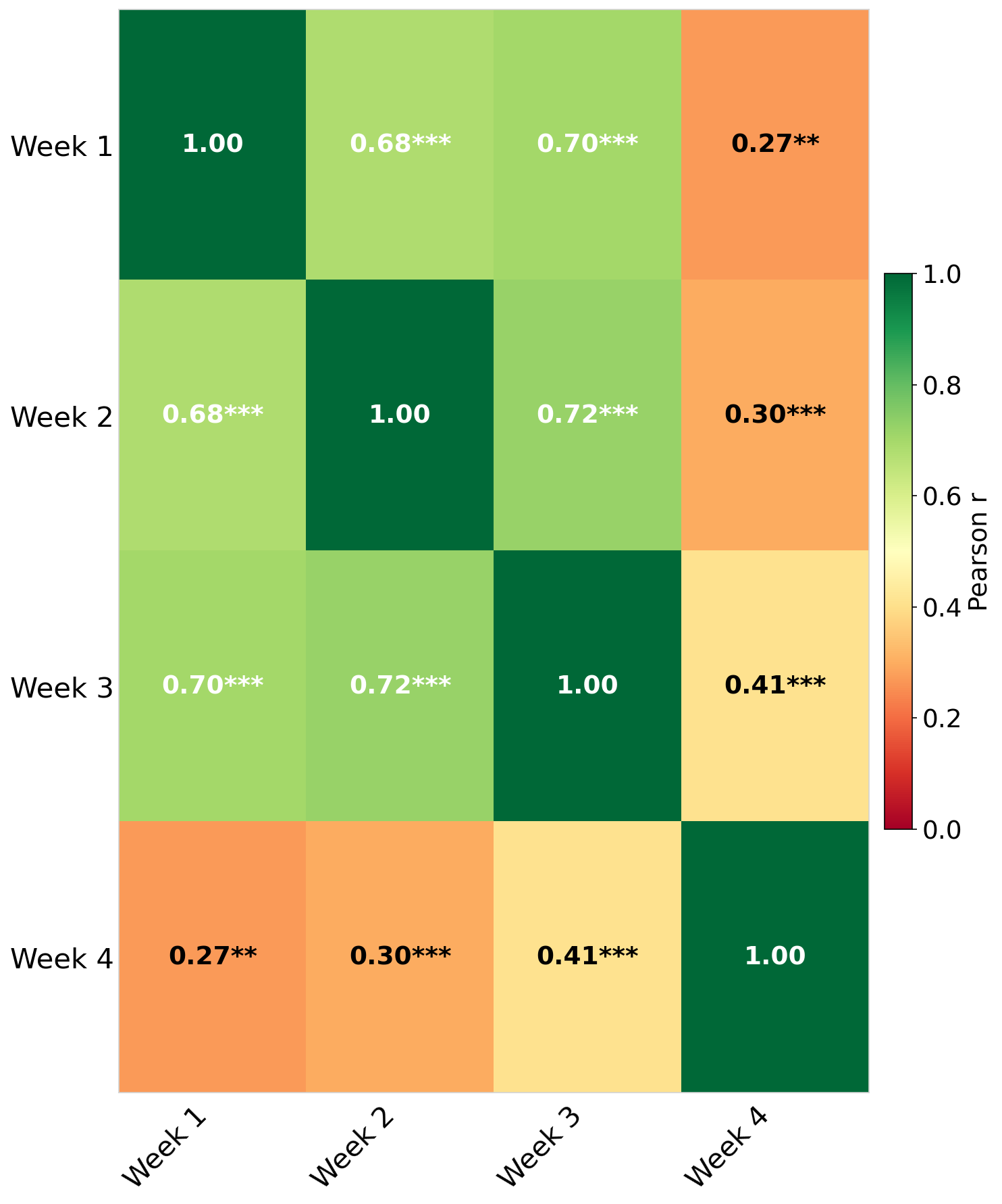}
    \caption{QAP heatmap comparison of semantic networks across events. ** indicates significant difference at the $p<0.01$ level, *** indicates significant difference at the $p<0.001$ level. }
    \label{fig:qap_heatmap}
\end{figure}

We also used the QAP (Quadratic Assignment Procedure) test on pairwise semantic networks to identify whether the pairs have correlated structure beyond chance. \autoref{fig:qap_heatmap} presents a heatmap of the pairwise difference of networks, and  \autoref{sec:appendix_semantic} presents the detailed statistical results. 
From Week 1 to Week 3, the semantic network have moderate to high correlations (Week 1-Week 2 $r=0.68$, Week 2-Week 3 $r=0.72$). When the narrative transitions from a personal competition to a legal framing at Week 4, the semantic network of Week 4 becomes largely different (Week 3-Week 4 $r=0.41$). 
Our semantic network analysis illustrate how critical exogenous events (e.g. court proceedings)reorganize semantic co-occurrence patterns, producing statistically significant shifts in discourse topology. In the scenario, the initial chatter was about the music and competition specifics, while the chatter changes to focus on the legal battle and the hostility between the competitors and the fans, which can be seen from the QAP differences.

The QAP correlations in Table 5 quantify the structural divergence between snapshots. Consecutive snapshots share moderate-to-high correlations (Week 1-Week 2: $r=0.68$; Week 2-Week 3: $r=0.72$; Week 3-Week 4: $r=0.41$), indicating that the semantic network evolves continuously but undergoes its sharpest structural break between Week 3-Week 4 — precisely when the narrative transitions from personal to legal framing. The Week 1–Week 4 correlation is the weakest of all pairs ($r=0.27, p<0.001$), confirming that the cumulative effect of the four events is a substantial reorganization of the discourse topology from its initial configuration. 

\subsection{Coordination Dynamics}
Agent-agent coordination dynamics are measured through the temporal synchronization of user actions around shared social media artifacts~\citep{magelinski2022synchronized,pacheco2021uncovering}. 
We calculate the Combined Synchronization Index ~\citep{ng2022online} of each user based on their social and semantic synchronization of actions. That is, over sliding time windows of five minutes, we measure the extent of to which a user tags other users with @mentions and uses the same sets of hashtags as other users. 
The five minute value is based on literature studies that show that beyond five minutes, there is a higher probability of coincidental rather than coordinated activity~\citep{graham2024coordination,magelinski2022synchronized,pacheco2021uncovering}.
We adapt the Combined Synchronization Index to handle cases of sparse coordination. For each agent pair $(u,v)$, we compute a normalized link weight $ LW(u,v)_{norm_k}$ within each coordination graph $G_k$. Examples of $G_k$ are semantic coordination graph, referral coordination graph or social coordination graph. The pairwise $CSI-Pair(u,v)$ is then the sum of these normalized weighs across all graph dimensions in which $(u,v)$ coordinate, divided by the total number of input graphs $T_nc$. This formulation ensures that the user pairs coordinating in only a small fraction of graphs are down-weighted relative to the user pairs that have persistent cross-dimensional coordination, and also reduces the negative average artifacts that arises from small number of coordinating graphs $n_c$ values that would arise from the original formulation. 

In our study, we calculated the semantic and social coordination indices. Semantic coordination arises around shared hashtags. For the semantic coordination graph $G_{h}=(V_h,E_h)$, nodes $v_i\in V_h$ represent agents that participate in semantic coordination. Edges $(e_1,e_2)\in E_h$ represent that the agents $v_1$ and $v_2$ had used a shared hashtag more than the 95th-percentile of users within five minute sliding time windows.
Social coordination arises around shared user mentions (@mentions). For the social coordination graph $G_{o}=(V_o,E_o)$, nodes $v_i\in V_o$ represent agents that participate in social coordination. Edges $(e_1,e_2)\in E_o$ represent that the agents $v_1$ and $v_2$ had used a shared user tag more than the 95th-percentile of users within five minute sliding time windows.

\begin{figure}
    \centering
    \includegraphics[width=1\linewidth]{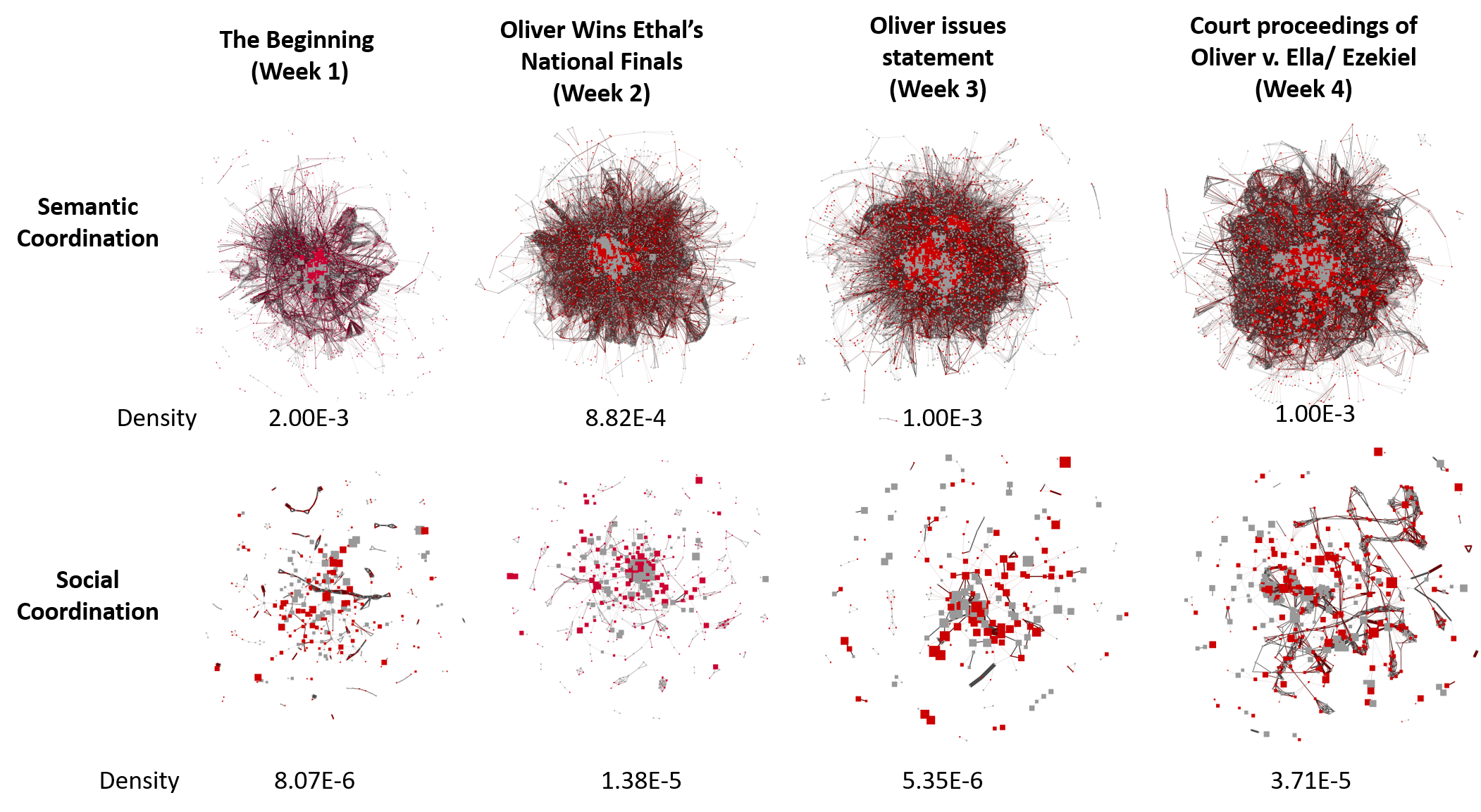}
    \caption{Coordination Networks. Nodes represent agents, which are sized by Combined Synchronization Index. Red nodes represent bot agents, grey nodes represent human agents. Edges between two nodes indicate that the two agents had coordinated with each other.}
    \label{fig:coordination_networks}
\end{figure}

The semantic and social coordination networks are presented in \autoref{fig:coordination_networks}. The two types networks reveal distinct structural signatures for the two measured synchronization mechanisms. Semantic coordination produces through the synchronization of hashtags a densely clustered network structure, where agents are clustered into a central core with some agents at the peripheral. In our study, the core of the networks are typically dominated by bot agents (red nodes), which exhibit significantly higher Combined Synchronization Index values than human agents. This reflects the behavioral scripting of bot archetypes to deploy coordinated hashtag campaigns, which is similar to their behavior in the real-world~\citep{khaund2021social,ng2022online}. Social coordination produced through the tagging of users through the @mention mechanism result in sparser graphs with chain-like cascade topology, reflecting how narratives propagate sequentially through tagged interaction chains~\citep{goel2016structural}. This structure resembles the directed trees in empirical retweet cascade networks. Human agents participate more prominently in social coordination while bot agents participate more in semantic coordination. This is consistent with evidence that human conversational behavior is more interpersonal while bot conversational behavior is more broadcast-oriented~\citep{ng2025global,ng2024exploratory}.

\begin{figure}
    \centering
    \includegraphics[width=1\linewidth]{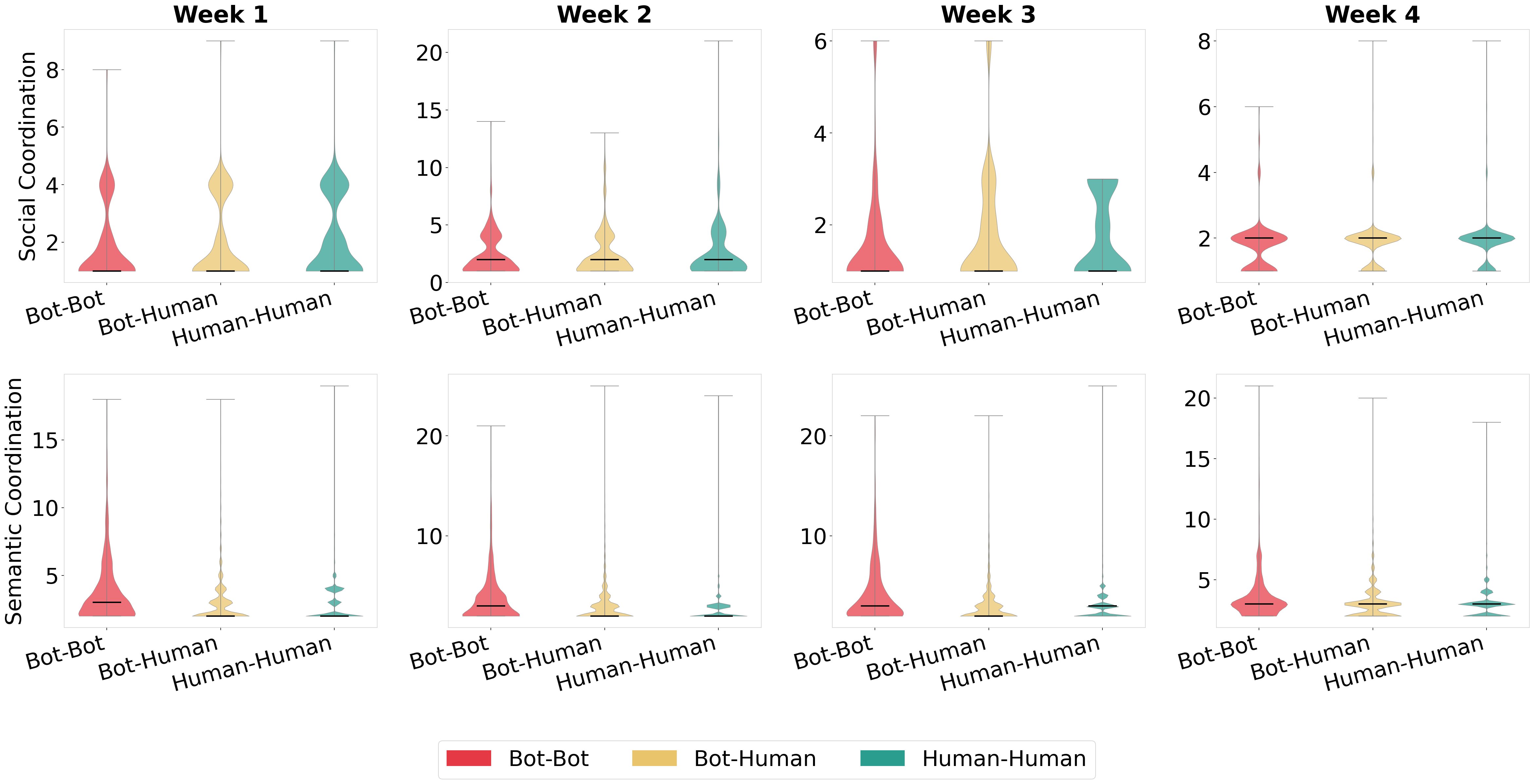}
    \caption{Comparison of coordination between dyad pairs of agent types. Coordination between two agents is measured through the temporal synchronization of user actions through social media artifacts (semantic coordination through shared hashtags, social coordination through shared @mentions). This graph compares dyad pairs of agent types, revealing that Bot-Bot and Bot-Human agent dyads have more coordination between them than the Human-Human dyad. }
    \label{fig:csi_dyad_combined}
\end{figure}

\autoref{fig:csi_dyad_combined} compares the coordination between dyad pairs (Bot-Bot / Bot-Human / Human-Human) across temporal snapshots, with the detailed quantitative statistics presented in \autoref{sec:appendix_coordination}. Bot participation in coordination is broadly proportional across snapshots, ranging from 45.4\% to 49.2\% of coordinating agents. For semantic coordination, Bot-Bot dyads consistently exhibit the highest mean edge weights across all snapshots (e.g. Day 17 $\mu_{bot}=4.11, \mu_{human}=1.74$), demonstrating that bot pairs engage in the most intensive coordinated hashtag co-deployment. In contrast, Human-Human dyads exhibit the largest raw edge counts but lower mean edge weights, reflecting diffuse but lower intensity coordination. For social coordination, the pattern reverses. Human-human dyads exhibit the largest total edge counts, while Bot-Bot dyads maintain higher mean coordination weights per dyad when they do engage. This reflects that @mention-based social interaction is predominantly a human conversational mechanism, given that there are a lot more Human-Human dyads than Bot-Bot dyads (e.g., Day 1: 96,078 Human-Human edges vs. 2,210 Bot-Bot edges).

Temporally, coordination volume peaks at Day 10 (Oliver Wins National Finals) across all dyad types. Bot-Human dyads reaching 1,246 edges in semantic coordination, before declining at Day 17 and surging again at Day 24 (Court Proceedings) with Bot-Human reaching 2,578 edges. This pattern of peaking around high salience events is consistent with observations that coordinated behavior intensifies in response to narrative triggers~\citep{graham2024coordination}. 
Finally, the converge of edge weight distributions across dyad types by Day 24 reflected from the tightening standard deviations of social and semantic coordination (\autoref{tab:edge_weight_semantic}, \autoref{tab:edge_weight_social}) suggests a coordination consolidation effect, where the heightened salience of the legal proceedings draws all agent types into uniformly intense narrative and interaction patterns.

\subsection{Influence Dynamics}
\begin{figure}
    \centering
    \includegraphics[width=\linewidth]{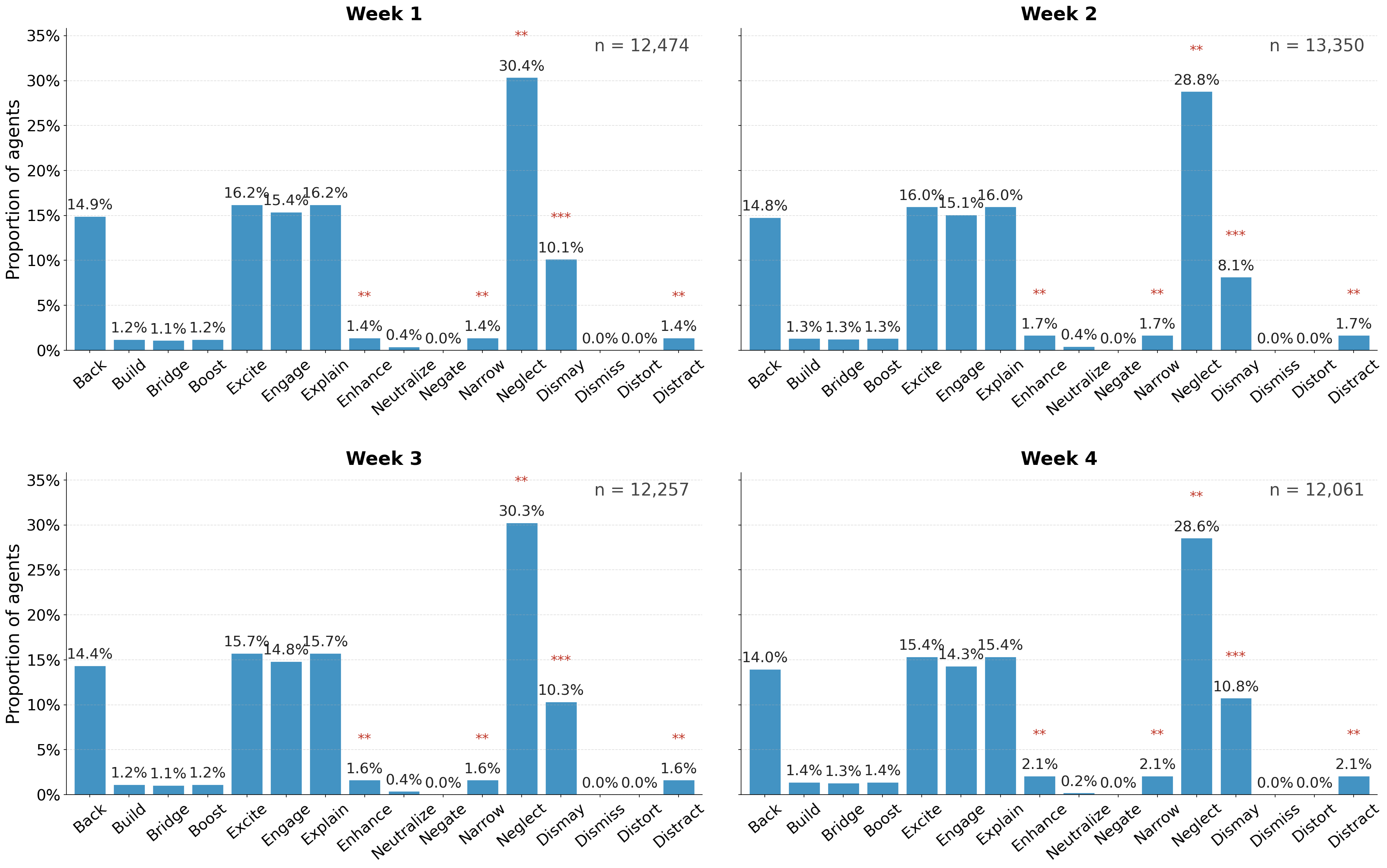}
    \caption{Influence dynamics over time measured by the BEND metrics. ** indicates $p<0.01$ and *** indicates $p<0.001$ for a Chi squares test of difference.}
    \label{fig:bend}
\end{figure}

Influence dynamics are examined through the BEND framework, which characterizes influence maneuvers along both narrative and network dimensions \citep{carley2026bend}. The BEND framework consists of sixteen adverse and affirmative maneuvers such as Back, Bridge, Negate or Dismiss, that describe how one agent attempts to influence another. This framework has been used to analyze influence dynamics in digital propaganda~\citep{marigliano2024analyzing}, cross-country information operations~\citep{alieva2022disinformation} and webgraphs~\cite{williams2025extending}, 
We quantitatively calculated the values of each influence maneuver using the ORA-Pro software~\citep{altman2020ora}. We present the proportion of agents that use each BEND maneuver across each time point in \autoref{fig:bend}.

The results in \autoref{fig:bend} reveal a stable baseline of affirmative maneuvers (Back, Build, Bridge, Boost, Excite) across four time points, which accounts about 14 to 16\% of agents each. This reflects the supportive discourse the agents participate with each other to reinforce their social groups.
The dominant maneuver across all time points are Neglect ($\sim$28-30\%), indicating that a plurality of agents consistently ignore or de-prioritize competing narratives rather than actively contesting them. 
Affirming complementary narratives while ignoring opposing narratives is a  structurally consequential influence strategy because it amplifies desired frames yet starves contradictory ones~\citep{slater2007reinforcing}. 

To test for changes in BEND maneuvers over time, for each of the 16 BEND metrics, we applied a Chi-square test of homogeneity. We used a 2x4 contingency table, where the rows represented agents that used the maneuver vs agents that did not use the maneuver, and the four columns represented each of the event dates. This test asks whether the proportion of agents using each BEND maneuver was the same or shifted over time. The resultant p-values were adjusted using the Benjamini-Hochberg (BH) false discover rate procedure. The results of the Chi-square test is presented in \autoref{tab:chi_square_results}.

The Chi square tests of homogeneity (\autoref{tab:chi_square_results}) identify five BEND maneuvers that shift significantly across the four event periods: Dismay ($\chi^2$=59.03, $p<0.001$), Distract ($\chi^2$=18.34, $p<0.01$), Narrow ($\chi^2$=18.34, $p<0.01$), Enhance ($\chi^2$=18.34, $p<0.01$) and Neglect ($\chi^2$=16.14, $p<0.01$). Dismay is an adverse maneuver that spreads fear, alarm or discouragement~\citep{carley2020social}, and shows the most dramatic temporal shift. Its proportion rises sharply as Oliver issues his statement and peaks at the Court Proceedings event. This is consistent with the narrative shift for the key actor Oliver, from a competitive framing at The Beginning (\textit{``I truly believe this is the moment Odrian pride shines brightest—winning the right to represent Odria at AuraSight in July 2030."}) to fan disappointment after the issue of the statement (\textit{``It's truly disappointing to see Oliver ignore the voice of Ethal. His silence and failure to withdraw that statement shows he forget where he came from"}) to an adversarial framing (\textit{``Once again, Ethal is overlooked and abandoned by Nareth. Our small nation faces constant neglect, betrayed by those who claim to support us.", ``Many Ethalians are concerned that Ezekiel and Ella lack the resources to fight both Odria and Oliver in courts. This situation isn't unique — similar cases across the globe have shown how powerful entities often dominate legal processes, leaving artists vulnerable. The overwhelming inequality in resource distribution indicates a systemic issue that affects Ethal’s entire artistic community."}). 

At the same time, there is a significant increase in Distract and Narrow maneuvers, which redirects attention towards tangential topics like Ella's and Ezikel's new single release (\textit{``don’t forget to keep streaming his single if you want to support", ``THE COUNTDOWN HAS BEGUN AND WEDNESDAY WILL BE UNFORGETTABLE!  REMEMBER, MY NEW SINGLE IS JUST THE BEGINNING OF SOMETHING EPIC!"}). As the legal conflict escalates, a subset of agents, particularly those involved in the public relations of our key artistes, converge on tactics designed to limit the visibility of exculpatory frames. Finally, the Enhance maneuvers, which amplifies the credibility of friendly narratives (\textit{``LET’S STAND UNITED AND BELIEVE IN OLIVER!", ``Let's celebrate artists who focus on talent, not stalking or petty jealousy! Remember, most fans are here for positive vibes and genuine connections!"}), shows a complementary temporal trajectory which increases slightly but significantly across the event horizon.

There is also a decrease in Excite and Explain maneuvers as the scenario progresses. This is because initial excitement about the countdown to the competition, the competition itself and the singers waned as time passes (\textit{``This is going to be a historic event!!", ``I CAN'T GET ENOUGH OF OLIVER'S SONGS!!"}). There is also less explanation being done by news agencies about the competition entrants and their songs (\textit{``Did you know that the first AuraSight took place in 1950 and has since become a cornerstone for regional cultural exchange?", ``AuraSight's influence remains significant, with media outlets like Regional Arts Review and government agencies like the Cultural Heritage Directorate recognizing its role in fostering regional unity and artistic innovation. "}).

The influence dynamics patterns in AuraSight mirror theoretical shifts in adversarial narrative management, where the objective is not really to win a single argument but to progressively constrain the discursive space available to opposing narratives~\citep{entman1993framing}. The GhostField architecture reproduces this sequencing organically and without any scripted directions. This suggests that such adversarial narrative management may not be detectable from a single isolated influence maneuver metric, but the temporal signature of the shift of maneuver portfolios over events in response to the narratives generated.

\begin{table}[ht]
\centering
\small
\begin{tabular}{lrrrlr}
\toprule
Maneuvers & $\chi^2$ & Raw $p$ & Adjusted $p$& Sig. & Direction of Change \\
\midrule
Dismay      & 59.03 & $9.47 \times 10^{-13}$ & $1.51 \times 10^{-11}$ & *** & DUU \\
Distract    & 18.34 & $3.75 \times 10^{-4}$  & $1.50 \times 10^{-3}$  & ** & UDU \\
Narrow      & 18.34 & $3.75 \times 10^{-4}$  & $2.00 \times 10^{-3}$  & ** & UDU \\
Enhance     & 18.34 & $3.75 \times 10^{-4}$  & $3.00 \times 10^{-3}$  & ** & UDU \\
Neglect     & 16.14 & $1.06 \times 10^{-3}$  & $3.40 \times 10^{-3}$  & ** & DUD \\
Neutralize  & 9.31  & $2.54 \times 10^{-2}$  & $6.78 \times 10^{-2}$  &  & - -D \\
Engage      & 5.84  & $1.20 \times 10^{-1}$  & $2.73 \times 10^{-1}$  &  & DUD \\
Back        & 5.69  & $1.28 \times 10^{-1}$  & $2.56 \times 10^{-1}$  &  & DDD \\
Bridge      & 4.50  & $2.12 \times 10^{-1}$  & $3.77 \times 10^{-1}$  &  & UDU \\
Build       & 3.63  & $3.04 \times 10^{-1}$  & $4.87 \times 10^{-1}$  &  & UDU \\
Boost       & 3.63  & $3.04 \times 10^{-1}$  & $4.42 \times 10^{-1}$  &  & UDU \\
Excite      & 3.57  & $3.11 \times 10^{-1}$  & $4.15 \times 10^{-1}$  &  & DDD \\
Explain     & 3.57  & $3.11 \times 10^{-1}$  & $3.83 \times 10^{-1}$  &  & DDD \\
Negate      & 0.00  & $1.00$                 & $1.00$                 &  & - - - \\
Dismiss     & 0.00  & $1.00$                 & $1.00$                 &  & - - - \\
Distort     & 0.00  & $1.00$                 & $1.00$                 &  & - - - \\
\bottomrule
\end{tabular}
\caption{Chi-square test of homogeneity results with Benjamini-Hochberg FDR correction. $p<0.01$ and *** indicates $p<0.001$. Direction of Change compares against the previous week, U: Up, D: Down, -: No change. }
\label{tab:chi_square_results}
\end{table}

\section{Validation through Stylized Facts}
\label{sec:stylized_facts}
The generated AuraSight scenario presents several stylized facts with regards to social dynamics, and these facts are actually comparable to real-life data. \autoref{tab:stylized_facts} lists several stylized facts and describe how they played out in the generated scenario and the real-life scenario. 

\begin{longtable}{p{3.5cm}p{4.5cm}p{5.0cm}}\\

\toprule
\textbf{Stylized Fact} & \textbf{AuraSight Scenario} & \textbf{Real-World Scenarios} \\
\midrule
\endfirsthead

\multicolumn{3}{c}%
{{\tablename\ \thetable{} -- continued from previous page}} \\
\toprule
\textbf{Stylized Fact} & \textbf{AuraSight Scenario} & \textbf{Real-World Scenarios} \\
\midrule
\endhead

\midrule
\multicolumn{3}{r}{{Continued on next page}} \\
\endfoot

\bottomrule
\endlastfoot

\multicolumn{3}{c}{\textbf{Agent Ego-Network Topologies}}\\
\midrule

Shifts in ego-network topologies change as a function of events
&
The ego-networks of key actors (Oliver, Ella, Ezekiel) expand and contract across the four time windows.
&
Ego-network structure shifts in response to crisis events. Central actors have denser networks around more salient moments \citep{ceria2022topological,cekini2026impact}.
\\

The centrality of actor nodes have a long-tail distribution
&
Oliver maintains a dominant eigenvector centrality (0.70) across all four weeks, while other actors occupy the tail with eigenvector centrality values below 0.1 for most of the scenario.
&
Degree and centrality in online networks follow heavy-tailed patterns, concentrating connections in a small fraction of nodes \citep{broido2019scale}.
\\

A small number of actors accumulate a disproportionate amount of attention and centrality
&
Oliver consistently holds high betweenness centrality (0.98) across all time windows, while most agents remain structurally peripheral.
&
High-profile accounts and news agents are disproportionately more central and retweeted more often than ordinary users \citep{deverna2024identifying}.
\\

\midrule
\multicolumn{3}{c}{\textbf{Semantic Networks}}\\
\midrule

Semantic networks for co-occurring words are denser than the social network formed from interactions
&
The semantic network of Week 4 is denser than that of Week 1 (730 edges vs.\ 181 edges).
&
COVID-19 Twitter discourse exhibits persistent topical hashtag clusters alongside temporally shifting ones as events unfold \citep{cruickshank2020characterizing}.
\\

Semantic networks grow while maintaining topical communities
&
The semantic network grows from Week 1 to Week 4 while maintaining high modularity.
&
Semantic networks grow by adding peripheral terms while preserving a densely connected conceptual core \citep{steyvers2005large}.
\\

Exogenous events trigger changes in semantic networks
&
The semantic network experiences a structural break between Weeks 3 and 4 (QAP $r=0.41$), which coincides with the transition from personal campaigning to the legal battle.
&
Political events, protests, and breaking news can trigger reorganization of hashtag co-occurrence networks \citep{huang2019hierarchical}.
\\

\midrule
\multicolumn{3}{c}{\textbf{Coordination Dynamics}}\\
\midrule

Bot coordination is more intensive than human coordination
&
Bot-Bot dyads exhibit the highest semantic coordination weights across all time periods.
&
Bot agents exhibit significantly higher coordinated hashtag deployment than human agents \citep{ng2023combined}
\\

Semantic coordination is stronger around events
&
Semantic coordination peaks at Week 2 and Week 4, which are the two highest-salience events, Oliver's win and the legal battle respectively.
&
Coordinated hashtag activity intensifies in response to politically salient events \citep{graham2024coordination}.
\\

\midrule
\multicolumn{3}{c}{\textbf{Influence Dynamics}}\\
\midrule

Influence strategies and narratives co-evolve
&
Five BEND maneuvers shift significantly across the four time windows.
&
Organizational influence strategies adapt in response to evolving narrative environments, and actors change their message framing as narratives develop \citep{vaara2016narratives,dahlen2009marketing}.
\\

The Back maneuver is always done by people who like an actor
&
Back maneuvers are consistently performed by pro-Oliver, pro-Ella, pro-Ezekiel actors throughout the scenario
&
In-group affirming behavior on social media is pre-dominantly performed by ideologically aligned users who amplify favorable narratives 
\citep{rathje2021out}
\\

The Negate maneuver is always done by people who dislike an actor
&
Negate shows zero temporal variation and is exclusively produced by anti-Oliver agents across all time windows
&
Out-group animosity drives counter-narrative engagement on social media
\citep{rathje2021out}
\\

Messages that contain Excite or Dismay are more likely to be retweeted
&
Dismay shows the largest temporal shift of all maneuvers ($\chi^2 = 59.03, p < 0.001$), rising sharply as the conflict escalates.
&
Emotionally activating content receives significantly higher retweet rates \citep{rathje2025psychology}.
\\
\bottomrule
\caption{Stylized facts and their observations in the AuraSight scenario and real-world settings.}
\label{tab:stylized_facts}
\end{longtable}

\section{Discussion}
\label{sec:discussion}
Through the AuraSight scenario, we demonstrate that in a networked and conversational environment, agents generate social dynamics of coordination and influence through the recursive interactions between network topology and narrative exchange. Here, network topology interactions are studied through ego-networks analysis, narrative exchange is studied through semantic network analysis, coordination dynamics is studied through agent-agent social and semantic coordination, and influence dynamics studied through the BEND maneuvers.

Ego-network evolution establishes the network topology dimension of this argument. Oliver's structural centrality was not prescribed but emerged from the recursive interaction process. His early narrative salience from his competition win attracted bot amplification, which expanded his network footprint, which in turn increased the reach of subsequent narratives associated with him. His eigenvector and betweenness centrality scores remain rather stable across all four snapshots (eigenvector: 0.703 to 0.697; betweenness: 0.988 to 0.981) even as the narrative environment shifts from celebratory to adversarial. Conversely, Ella and Ezekiel increase in structural salience (as measured by their eigenvector and betweenness centrality of their ego-networks) only when the narrative frame shifts to implicate them directly, demonstrating that network topology is not a static backdrop against which narratives play out but is itself continuously reshaped by narrative events~\citep{granovetter1985economic,cauteruccio2026structure,carley2009etiology}.

The semantic network results establish the narrative exchange dimension. The discourse topology reorganizes in structurally coherent ways around each exogenous event, as reflected in the progressive decline in QAP correlations with temporal distance (Week 1-Week 4: $r=0.27$) and the sharpest structural break occurring between Week 3 and Week 4 ($r=0.41$). This narrative reorganization is not only a content shift, but also has structural consequences. As the semantic network transitions from competition framing to a legal framing, the hashtag communities that constitute the discourse backbone reconfigure, which channels agent intreractions differently, affecting the extent of coordination and the types of agents present (e.g. Oliver has an expanded ego-network with a lot more bots at Week 3, which collapsed into a core-periphery structure with lesser bots by Week 4). Narrative exchange thus can also be a force that shapes interaction topologies rather than merely a record of agent-agent conversations~\citep{diesner2005revealing,danowski1993network,bail2016combining}. Our results contribute to a growing body of work that demonstrate that social dynamics in a networked environments are not reducible to individual agent cognition, but emerge from the recursive interplay of network topology and content-level dynamics~\citep{carley2009etiology,weng2012competition,vosoughi2018spread}.

The coordination analysis and influence maneuver results show what happens when the two dimensions operate together. Coordination is structurally bifurcated along the topology-narrative divide. 
Semantic coordination networks are substantially denser than social coordination networks across all four time windows, reflecting the broadcast nature of hashtag deployment compared to the targeted dyad-level nature of social coordination.
Bot agents are concentrated at the structural core of semantic coordination networks, while human agents dominate social coordination networks, depicting the difference in strategies through broadcast vs interpersonal interactions.
Influence maneuver shifts are also a product of the network-narrative recursive process. As the event progresses, the semantic network re-organizes around more adversarial frames (\textit{``\#weknowthetruth, \#innocentoli, \#odrianethal}), and the proportion of agents deploying the Dismay, Distract, and Narrow influence maneuvers rises significantly (all $p<0.01$). Since there were no instruction injects to any agents to shift strategy through the simulation, this emergence in influence maneuvers is indeed a product of the narrative environment. Coordination and influence dynamics are indeed emergent properties of the network-narrative interplay in the simulation. Agents coordinate because the network position brings them into contact with the same narratives, to which they attach through preferential attachment or leader-following mechanisms; and influence maneuvers shift over time because the narrative environment restructures the topology of attention. 

This work sits at the intersection of social cybersecurity and multi-agent social systems research, offering a computational lens through which the entanglement of network structure and online discourse dynamics can be controlled in a realistic environment~\citep{carley2020social,ng2026social}.
While our study of LLM-Enabled Agent Based Network-Dynamic models offers a pathway towards realistic experimental social simulations, there are still some limitations that warrants future work. First, there are several assumptions made in the architecture. For example, agent activation is currently approximated by probabilistic rules like Bernoulli processes and fixed intervals. While these abstractions estimate the online rhythm of posting activities, they omit micro-level cognitive factors like attention, fatigue recovery and feedback adaption. Future work involves refining this persona generation system to incorporate more behavioral realism into the agents for more accurate simulation. Second, our study of social dynamics treat exogenous events equally, omitting the fact that these events may have different effects on the network because they affect each individual type differently. Future work involves incorporating finer grained analysis to such temporal dynamics for a more realistic simulation of information propagation in a network.

\section{Conclusions}
\label{sec:conclusions}
This paper presented AuraSight, a social simulation scenario generated using the GhostField architecture. 314,244 agents exchanged messages across four weeks of a fictional, real-life inspired online event. The simulation operationalized both networked interaction and linguistic discourse within a conversational environment. This enables the joint study of network and linguistic interactions that previous studies had to treat separately. The AuraSight scenario generated social dynamics around ego-network evolution, semantic discourse organization, coordinated behavior patterns and influence maneuver deployment, that are theoretically coherent with literature through validation of a set of stylized facts.
Finally, the social simulation architecture and social dynamics analysis methodology provides a replicable and extensible testbed for studying the social dynamics of cyber social agents under controlled experimental conditions. 

Our work bridges structural modeling and discourse generation, and have implications for how online social dynamics are theorized and measured. Models that treat social dynamics as a property of agent attributes (e.g., follower count, posting frequency, bot likelihood), will underestimate the structural dimension of agent interactions. Conversely, purely structural models that ignore narrative content will miss the content-driven changes in network topology. A complete account of online social dynamics requires modeling both simultaneously, and our study offers a computational framework for doing so. 


\begin{acks}
This material is based upon work supported by the Scalable Technologies for Social Cybersecurity, U.S. Army (W911NF20D0002), the Threat Assessment, Office of Naval Research (N000142412414), and Community Assessment, Office of Naval Research (N000142412568), and ACTAINS, Office of Naval Research (N000142612511). The views and conclusions contained in this document are those of the authors and should not be interpreted as representing official policies, either expressed or implied by the Office of Naval Research, U.S. Army or the U.S. government.
\end{acks}

\bibliographystyle{ACM-Reference-Format}
\bibliography{acmart}

\appendix

\section{Agent Ego-Network Topologies Results}
\label{sec:appendix_egonetwork}
\begin{figure}[h]
    \centering
    \includegraphics[width=\linewidth]{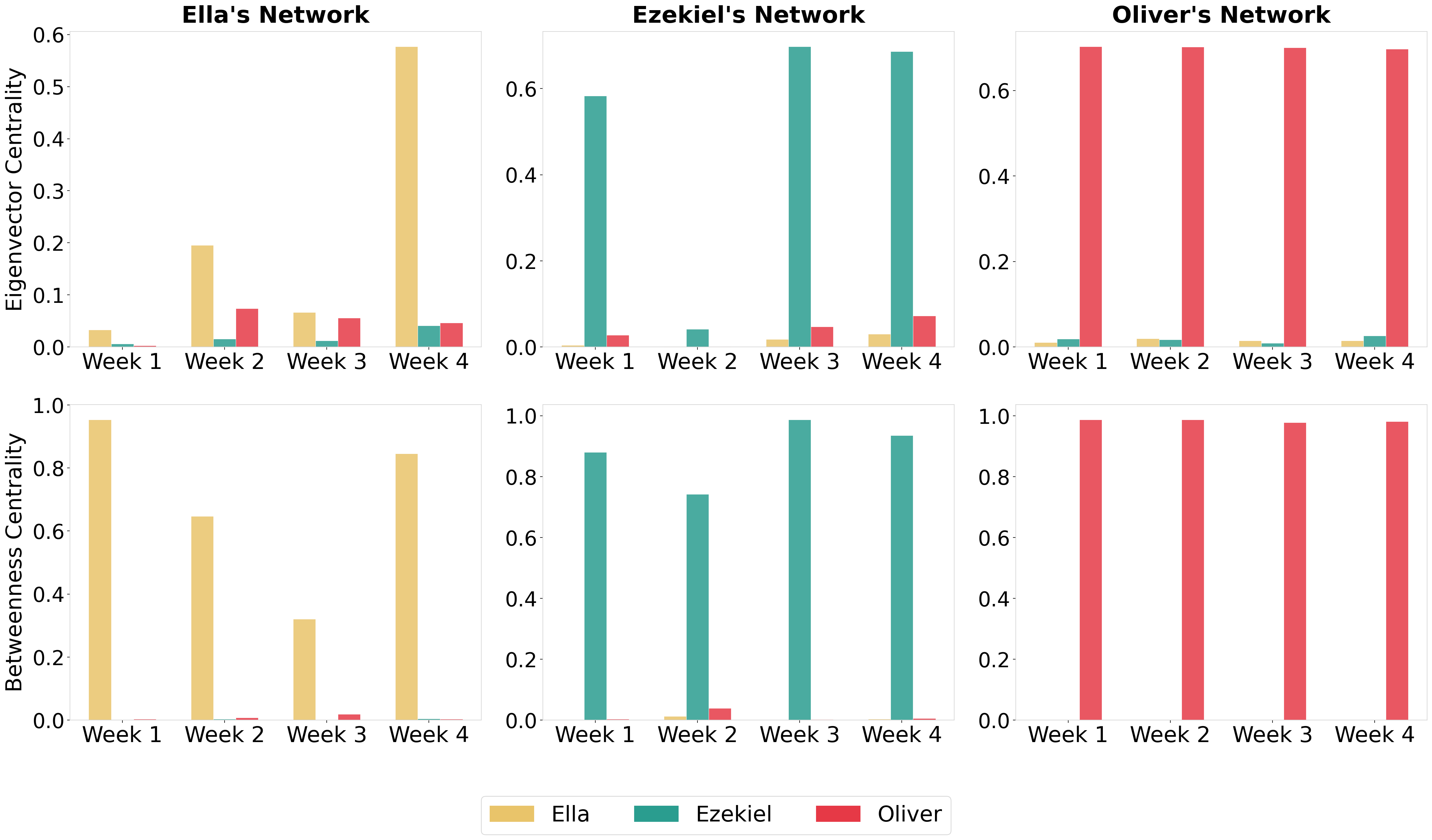}
    \caption{Centrality values of the main characters Oliver, Ella and Ezekiel across the event timestamps.}
    \label{fig:main_character_centrality}
\end{figure}

\autoref{fig:main_character_centrality} presents the centrality values of the main characters Oliver, Ella and Ezekiel across event snapshots, and the detailed scores are in \autoref{tab:actor_centrality_scores}. \autoref{tab:centrality_correlations} presents the correlations of networks calculated using the QAP procedures. 

\begin{table}[t]
\centering
\small
\begin{tabular}{lccc}
\toprule
Snapshot & Ella & Ezekiel & Oliver \\
\midrule
\multicolumn{4}{l}{\textbf{Eigenvector Score}} \\
\midrule
The Beginning (Week 1) & 0.033 & 0.582 & 0.70 \\
Oliver Wins Ethal's National Finals (Week 2) & 0.20 & 0.04 & 0.70 \\
Oliver Issues Statement (Week 3) & 0.067 & 0.70 & 0.70 \\
Court Proceedings (Week 4) & 0.58 & 0.69 & 0.70 \\
\midrule
\multicolumn{4}{l}{\textbf{Betweenness Score}} \\
\midrule
The Beginning (Week 1) & 0.95 & 0.88 & 0.98 \\
Oliver Wins Ethal's National Finals (Week 2) & 0.65 & 0.741 & 0.98 \\
Oliver Issues Statement (Week 3) & 0.32 & 0.98 & 0.98 \\
Court Proceedings (Week 4) & 0.85 & 0.935 & 0.98 \\
\bottomrule
\end{tabular}
\caption{Eigenvector centrality and betweenness centrality scores of the three principal actors across network snapshots.}
\label{tab:actor_centrality_scores}
\end{table}

\begin{table}[ht]
\centering
\small
\begin{tabular}{lrrrll}
\toprule
Pair & Shared $N$ & Pearson $r$ & $p$-value & Sig. & Interpretation \\
\midrule
Week 1 vs Week 2 & 45 & 0.43 & $p<0.001$ & *** & Moderately similar \\
Week 1 vs Week 3 & 45 & 0.37 & $p<0.001$ & *** & Weakly similar \\
Week 1 vs Week 4  & 45 & 0.31 & $p<0.001$ & *** & Weakly similar \\
Week 2 vs Week 3 & 44 & 0.34 & $p<0.001$ & *** & Weakly similar \\
Week 2 vs Week 4  & 46 & 0.40 & $p<0.001$ & *** & Moderately similar \\
Week 3 vs Week 4 & 49 & 0.36 & $p<0.001$ & *** & Weakly similar \\
\bottomrule
\end{tabular}
\caption{Pearson correlations of actor centrality rankings across network snapshots. Higher correlations indicate greater stability in the relative prominence of actors between snapshots.}
\label{tab:centrality_correlations}
\end{table}

\section{Semantic Network Analysis Results}
\label{sec:appendix_semantic}
\autoref{tab:network_snapshots} presents the network statistics across key snapshots of the semantic analysis networks, and \autoref{tab:network_correlations} presents the correlation of node centrality rankings across semantic network snapshots.

\begin{table}[t]
\centering
\small
\begin{tabular}{p{3cm}rrrrrr}
\toprule
Snapshot & Nodes & Edges & Communities & Modularity & Components & Largest Component \\
\midrule
Week 1 & 85 & 238 & 9 & 0.74 & 6 & 44 \\
Week 2 & 141 & 491 & 15 & 0.78 & 4 & 126 \\
Week 3 & 149 & 521 & 14 & 0.75 & 6 & 125 \\
Week 4 & 181 & 730 & 13 & 0.78 & 4 & 168 \\
\bottomrule
\end{tabular}
\caption{Network statistics across key snapshots of the semantic analysis network.}
\label{tab:network_snapshots}
\end{table}

\begin{table}[t]
\centering
\small
\begin{tabular}{lrrrrll}
\toprule
Pair & Shared Nodes & Pearson $r$ & $p$-value & Sig. & Interpretation \\
\midrule
Week 1 vs Week 2 & 73  & 0.68 & $p<0.001$ & *** & Moderately similar \\
Week 1 vs Week 3 & 63  & 0.70 & $p<0.001$ & *** & Moderately similar \\
Week 1 vs Week 4 & 63  & 0.27 & 0.001      & **  & Weakly similar \\
Week 2 vs Week 3 & 84  & 0.72 & $p<0.001$ & *** & Highly similar \\
Week 2 vs Week 4  & 83  & 0.30 & $p<0.001$ & *** & Weakly similar \\
Week 3 vs Week 4  & 106 & 0.41 & $p<0.001$ & *** & Moderately similar \\
\bottomrule
\end{tabular}
\caption{Correlation of node centrality rankings across semantic network snapshots.}
\label{tab:network_correlations}
\end{table}

\section{Coordination Analysis Results}
\label{sec:appendix_coordination}
\autoref{tab:agent_composition} details the statistics of agent coordination composition. \autoref{tab:edge_weight_semantic} details the statistics of dyad pairs for semantic coordination, and \autoref{tab:edge_weight_social} details the statistics of dyad pairs for social coordination.

\begin{table}[t]
\centering
\small
\begin{tabular}{lrrrr}
\toprule
Snapshot & Total Agents & Bots & Humans & \% Bots \\
\midrule
Week 1 & 431 & 197 & 234 & 45.7\% \\
Week 2 & 498 & 245 & 253 & 49.2\% \\
Week 3 & 256 & 125 & 131 & 48.8\% \\
Week 4 & 641 & 291 & 350 & 45.4\% \\
\bottomrule
\end{tabular}
\caption{Agent coordination composition across network snapshots.}
\label{tab:agent_composition}
\end{table}

\begin{table}[t]
\centering
\small
\begin{tabular}{p{4.5cm}lrrrr}
\toprule
Snapshot & Dyad & $N$ Edges & Mean & Median & SD \\
\midrule
Week 1 & Bot-Bot & 234 & 1.76 & 1.000 & 1.28 \\
Week 1 & Bot-Human & 614 & 1.98 & 1.000 & 1.411 \\
Week 1 & Human-Human & 408 & 2.08 & 1.000 & 1.37 \\
\midrule
Week 2 & Bot-Bot & 584 & 2.36 & 2.000 & 1.79 \\
Week 2 & Bot-Human & 1246 & 2.44 & 2.000 & 2.03 \\
Week 2 & Human-Human & 634 & 2.51 & 2.00 & 2.33 \\
\midrule
Week 3 & Bot-Bot & 232 & 1.53 & 1.00 & 1.13 \\
Week 3 & Bot-Human & 400 & 1.755 & 1.00 & 1.27 \\
Week 3 & Human-Human & 172 & 1.74 & 1.00 & 0.89 \\
\midrule
Week 4 & Bot-Bot & 1032 & 1.74 & 2.00 & 0.78 \\
Week 4 & Bot-Human & 2578 & 1.80 & 2.00 & 0.71 \\
Week 4 & Human-Human & 1788 & 1.84 & 2.00 & 0.67 \\
\bottomrule
\end{tabular}
\caption{Descriptive statistics of edge weights by dyad type across network snapshots for \textbf{Semantic Coordination}. Mean, median, and standard deviation (SD) are computed over weighted interactions between bot--bot, bot--human, and human--human pairs.}
\label{tab:edge_weight_semantic}
\end{table}

\begin{table}[t]
\centering
\small
\begin{tabular}{p{5cm}lrrrr}
\toprule
Snapshot & Dyad & $N$ Edges & Mean & Median & SD \\
\midrule
Week 1 & Bot-Bot & 2210 & 3.75 & 3.00 & 2.33 \\
Week 1 & Bot-Human & 21096 & 2.98 & 2.00 & 1.66 \\
Week 1 & Human-Human & 96078 & 2.82 & 2.00 & 1.20 \\
\midrule
Week 2 & Bot-Bot & 273 & 3.78 & 3.000 & 2.430 \\
Week 2 & Bot-Human & 16908 & 3.118 & 2.000 & 1.929 \\
Week 2& Human-Human & 53814 & 2.646 & 2.000 & 1.111 \\
\midrule
Week 3 & Bot-Bot & 1202 & 4.11 & 3.00 & 2.970 \\
Week 3 & Bot-Human & 12414 & 3.13 & 2.00 & 1.986 \\
Week 3 & Human-Human & 53644 & 2.85 & 3.00 & 1.247 \\
\midrule
Week 4 & Bot-Bot & 2052 & 3.51 & 3.00 & 1.99 \\
Week 4 & Bot-Human & 15632 & 3.10 & 3.00 & 1.48 \\
Week 4 & Human-Human & 43668 & 2.94 & 3.0008 & 1.09 \\
\bottomrule
\end{tabular}
\caption{Descriptive statistics of interaction frequencies by dyad type across network snapshots for \textbf{Social Coordination}. Mean, median, and standard deviation (SD) summarize the number of interactions exchanged between bot--bot, bot--human, and human--human pairs.}
\label{tab:edge_weight_social}
\end{table}

\end{document}